# PatentSBERTa: A Deep NLP based Hybrid Model for Patent Distance and Classification using Augmented SBERT

-- Preliminary Draft, Work in Progress --

**Hamid Bekamiri**\*, **Daniel S. Hain, Roman Jurowetzki**

Aalborg University Business School, Denmark

## Abstract
This study provides an efficient approach for using text data to calculate patent-to-patent (p2p) technological similarity, and presents a hybrid framework for leveraging the resulting p2p similarity for applications such as semantic search and automated patent classification. We create embeddings using Sentence-BERT (SBERT) based on patent claims. To further increase the patent embedding quality, we use transformer models based on SBERT and RoBERT, and apply the augmented approach for fine-tuning SBERT by in-domain supervised patent claims data. We leverage SBERTs efficiency in creating embedding distance measures to map p2p similarity in large sets of patent data. We deploy our framework for classification with a simple Nearest Neighbors (KNN) model that predicts Cooperative Patent Classification (CPC) of a patent based on the class assignment of the K patents with the highest p2p similarity. We thereby validate that the p2p similarity captures their technological features in terms of CPC overlap, and at the same demonstrate the usefulness of this approach for automatic patent classification based on text data. Furthermore, the presented classification framework is simple and the results easy to interpret and evaluate by end-users. In the out-of-sample model validation, we are able to perform a multi-label prediction of all assigned CPC classes on the subclass (663) level on 1,492,294 patents with an accuracy of 54% and F1 score > 66%, which suggests that our model outperforms the current state-of-the-art in text-based multi-label and multi-class patent classification. We furthermore discuss the applicability of the presented framework for semantic IP search, patent landscaping, and technology intelligence. We finally point towards a future research agenda for leveraging multi-source patent embeddings, their appropriateness across applications, as well as to improve and validate patent embeddings by creating domain-expert curated Semantic Textual Similarity (STS) benchmark datasets.

---

# 1. Introduction

Technological distance, or the extent to which a set of patents represent the same or different types of technologies, is a key characteristic in being able to analyze and visualize innovative opportunities (Breschi et al., 2003). Such measures are in general extremely useful for technology mapping, analysis, and forecasting, for instance to create technology spaces (e.g. Alstott et al., 2017; Kogler et al., 2013), predicting technology convergence (e.g. San Kim and Sohn, 2020), early detection of disruptive technology (e.g. Zhou et al., 2020) and assessing patent quality and novelty (e.g. Artset al., 2018).

While traditionally such patent-to-patent (p2p) similarity measures have been created based on metadata such as keywords, technology classifications, or citations, recently semantic-based methods from the field of natural language processing (NLP) that utilize textual data such as abstracts, claims of full-text have increasingly gained popularity. Based on volumes of patents dataset, the deep learning approach had better results than traditional keyword-based approaches. Particularly, text embedding techniques, which represent textual data as a numeric vector while preserving its meaning, have proven useful to create semantically informed p2p similarity measures.

Transformers as new semantic-based methods have made the analysis of semantic-level information and retrieval from massive unstructured textual data for practical implementation possible (Sarica et al., 2020). These models are the current state-of-the-art way of dealing with sequences. Transformers get around the problem of lack of memory in RNNs and LSTMs by perceiving entire sequences simultaneously. As unsupervised language models, transformers are characterized by current state-of-the-art performance and effectiveness for downstream tasks such as Semantic similarity and classification. Recently, the first adaptations and applications of transformer models for patent classification (eg. PatentBERT by Lee & Hsiang, 2020) have emerged. While such applications have led to state-of-the-art performance in text-based patent classification, they also face some limitations.

First, up to now applied transformer models such as BERT are computationally expensive and therefore require a substantial computing infrastructure to be suitable and practical for large-scale deployment. Furthermore, their architecture is geared towards downstream prediction taste based on the on-the-fly created embeddings. This makes it particularly expensive to use such an architecture to calculate p2p similarity measures. We solve this problem by using a SBERT model. SBERT is a modification of the pre-trained BERT network that uses siamese and triplet network structures to derive semantically meaningful sentence embeddings that can be compared using cosine similarity. This reduces the effort for finding the most similar pair in a collection of 10,000 sentences from 65 hours with BERT / RoBERTa to about 5 seconds with SBERT while maintaining the accuracy from BERT (Reimers and Gurevych, 2019).

Second, pre-trained models based on generic text data face limitations when applied to domain-specific jargon. In patent texts, technical jargon, the use of legal language and attention to intellectual property rights here represent a challenge that can significantly affect the accuracy of patent analysis applications (QI et al., 2020). This effect can be mitigated by fine-tuning these models with in-domain examples labeled by domain experts. However, the labeled data in most downstream fields is sparse or not available, and the manual process of data labeling is costly. Although the power of BERT-based models pre-trained on large well-formed text corpora was demonstrated, the performance of pre-train them on in-domain is still debatable (Pota et al., 2021). Therefore, other ways to label data like the semi-supervised approach can be interesting for increasing model performance. In (Thakur et al., 2020), they proposed an efficient solution as a data augmentation strategy, where they used the cross-encoder like BERT, RoBERTa, etc., to label a more extensive set of input pairs to augment the training data for the bi-encoder. They showed that this approach achieved up to 6 points for in-domain and up to 37 points for domain adaptation tasks than the original bi-encoder performance.

Third, when deployed for patent technology classification, the multi-label as well as multi-class nature of the prediction task represents a significant challenge for researchers. In multi-label classification, each instance in the training set is associated with a set of labels, and the task is to predict a label set for each unseen instance. The problem of multi-label classification (MLC) is still one of the most significant research areas. This field has recently attracted increasing research due to its widespread applications in new technologies such as text classification and social network analysis (Zhen-Wu et al., 2020).

By addressing these issues with the framework presented in this paper, our contribution to the body of research- as well as to the range of applications-in patent analytics is at least fourfold. First, we provide a fast and efficient framework for utilizing transformer models to calculate p2p similarity measures. This enables fundamental application for semantic search, patent landscaping, technology mapping, and patent quality evaluation. Second, by applying augmented SBER, we present a practical workflow for fine-tuning transformer models to domain-specific language to be found in textual patent data. While this approach can be further improved by providing a small amount of domain-expert labeled examples, we demonstrate the usefulness of augmented approaches even without such labels in a purely self-supervised workflow. Third, we propose a hybrid model based on transformers and traditional ML models that outperforms the current state-of-the-art in text-based multi-label and multi-class patent classification. Fourth, using a simple KNN model for patent classification, we provide an easy and intuitive way to inspect, understand, and explain model predictions.

The remainder of the paper is structured as follows. In section 2, we review related work on NLP-based patent analytics, p2p similarity measures, and automated patent classification. Section 3 introduces the patent dataset used as well as its properties and methodological choices with respect to the patent information used in our framework, namely patent claim text and patent CPC classification. Section 4 introduces our SBERT based hybrid framework for patent embedding, p2p similarity calculation, and automated patent classification. The following section 5 presents our results and provides a validation exercise for patent classification. Section 6 concludes and points towards open questions and promising areas of future research.

## 2. Related work

Generally, different semantic similarity approaches can be categorized into four groups including keyword-based approaches, the analysis of the SAO structure, ontology-based analysis, as well as ML-based approaches (Hain et al., 2021).

Keyword-based methods are based on keyword frequency and co-occurrence measures. In this approach, raw (Arts et al., 2018) or TF-IDF weighted (Arts et al., 2020) methods can be used for creating similarity measures.

In the analysis of the SAO structure, a triple structure has been extracted from a text corpus that included Subjects, objects, and Actions. Subjects and Objects are related to the topic and Actions are verbs that represent the relationship between those terms and phrases. In this way, SAO structures can express semantic information better than Keyword-based methods (Yang et al., 2017).

In the ontology-based analysis approach, the concepts and related relations were defined for a specific domain. Based on this structure then a semantic annotation on patent texts is performed (Hain et al., 2021). Examples are the analysis system proposed by Taduri et al. (2011) and Soo et al. (2006).

While machine-learning-based approaches for text classification have been around since the 1990s (Hayes and Weinstein, 1990; Newman, 1998), they have only recently found growing attention in analyzing the semantic similarity of patents. The related methods in this approach are able to map the complex relationships of texts and provide high accuracy compared to other approaches (Hain et al., 2021).

According to the importance of the machine learning approach, lately considerable scientific studies are being conducted to develop these methods to determine how to estimate semantic similarity and patent classification. Most of the studies on patent classification are based on Deep learning and Transformers, but due to the lack of access to the standard datasets and the same evaluation structure, a fair review of the literature is very difficult. Whereas, the existence of different approaches in the classification like IPC and CPC and diversity of information, both structured and unstructured such as Date, Claim, Abstract, and Description have added to this difficulty (Li et al., 2018).

Chen and Chang (2012) proposed the hybrid approach as a three-phase categorization method that includes SVM, K-means, and KNN algorithms. In this approach, they used TF–IDF to select discriminative terms to classify 9062 patent documents as test datasets (12,042 for training) and the best result for the model was 53.25% accuracy at the sub-group level (Chen and Chang, 2012).

Tran and Kavuluru (2017) explore text data and machine learning-based classification in the context of the CPC system. Such exercises build on earlier work of automated patent classification for IPC classes (Fall et al., 2003). Grawe et al. (2017) introduce deep learning and word embedding methods for text-based patent classification.

Hepburn (2018) proposed the idea of using the transformers model for patent classification. Using a Support Vector Machine (SVM) and Universal Language model with Fine-tuning (ULMFiT), an F1 score of 78.4% for the prediction of IPC (section level, 8 labels) was achieved.

Li et al. (2018) proposed DeepPatent as a deep learning algorithm for patent classification based on convolutional neural networks (CNN) and word vector embedding of patent title and abstract. Their results in F1 were about 43%. Their study showed that DeepPatent achieved 73.88% classification precision by automatic feature extraction, which performed better than all existing algorithms that used the same information for training. They evaluated their model on USPTO-2M with 2,000,147 records after data cleaning of 2,679,443 USA raw utility patent documents in 637 categories at the IPC subclass level (Li et al., 2018).

Lee and Hsiang (2020) proposed PatentBERT as a transformers model based on the BERT-Base pre-trained model for patent classification. They focused on fine-tuning the BERT model for patent classification and used a large dataset USPTO of over two million patent claims at the CPC subclass level. Lee and Hsiang showed that patent claims alone are sufficient for the patent classification task, and they used patent claims to replace title and abstract which the F1 value approximately was equal in these different approaches. Their results of F1 Top 5 for IPC+Title+Abstract was 44.75% (Lee & Hsiang, 2020). Although In this approach they proposed a model with appropriate accuracy, they did not provide similarity measures between patents or suggest a workflow of how to do so (Hain et al., 2021). In this study, PatentBERT and DeepPatent are used as the baseline to benchmark.

## 3. Data

This section reviews the structure of the patent dataset and detailed patent classification approaches. First, The patent datasource in this study is brought up and in the next section the textual data of the patent document and classification approaches for patent will be discussed.

### Dataset Description

In this study, we used the PatentsView dataset. PatentsView is a patent data analysis platform designed to enhance the usefulness and transparency of US patent data and supported by the Office of Senior Economists at the US Patent and Trademark Office (USPTO). The PatentsView platform is built on a regularly updated database connecting inventors, organizations, locations, and the overall patent activity longitudinally (USPTO, 2020). In addition to traditional bibliographic data, it also contains textual patent data such as title, abstract, and claims, making it particularly valuable for deep NLP-based patent analytics. This dataset has been used for similar tasks in previous research, such as

DeepPatent (Li et al., 2018) and PatentBERT (Lee & Hsiang, 2020), making it a suitable dataset to benchmark our framework against previous work and the current state-of-the-art in deep NLP based patent classification.

In this study we utilized all patents between 2013-2017 which have at least one claim on the Google patent public datasets on Bigquery released in 2017. The number of records available in this study was 1,492,294 patents and we used 8 percent of the patents as the test dataset for evaluating the model. It is worth mentioning that we removed all duplicate records in patent id and claim text.

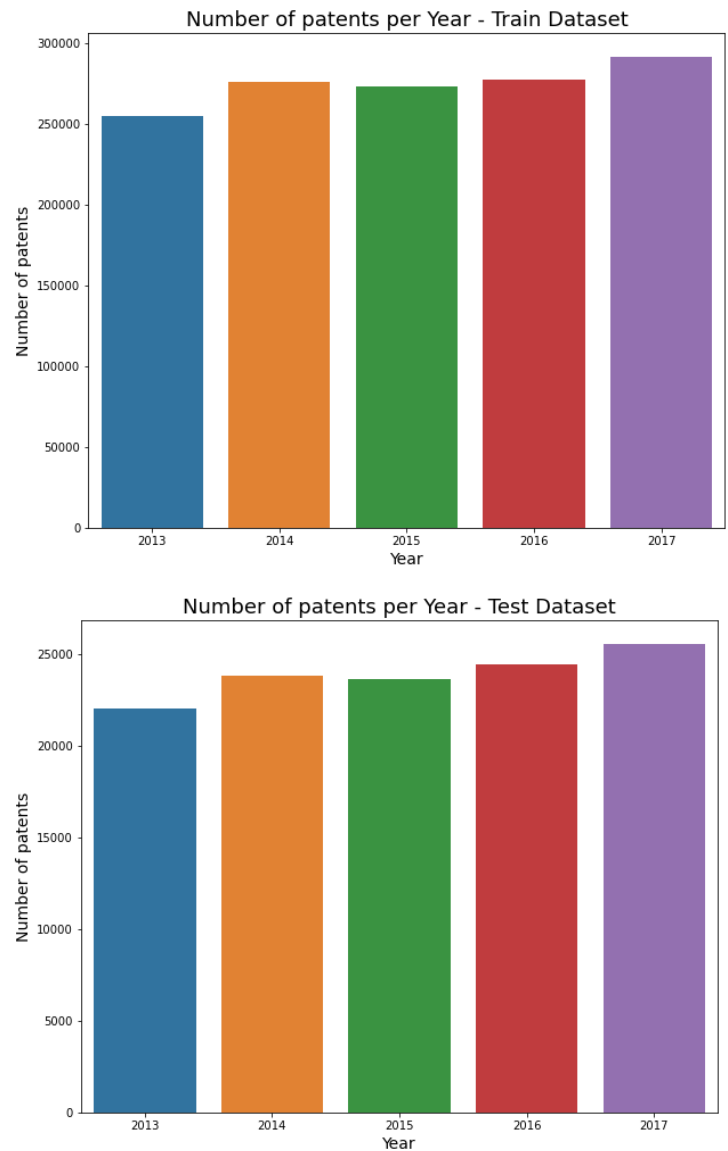

Figure 1: Number of patents per year

## Textual Data: Patent Claims

Among the options to use textual data to create patent embeddings (title, abstract, claims), we in this study utilized patent claim data. Claims are more important than ever in preparing patents. Claims are considered as the initial framework for preparing patent documents and other documents are collected and prepared based on them. Therefore, they contain more comprehensive and accurate information rather than other patent documents (Lee & Hsiang, 2020). Claims have a hierarchical structure that the first claim is considered as the backbone of this structure. Therefore, in this study, we used only the first claim and in the future studies; we want to combine all claims based on their tree structure (QI et al., 2020) for calculating semantic similarity and predicting multi-label classification. In the research sample, the mean value of the claims' number was 17. Table 1 provides an example of the hierarchical structure of the claims.

| Patent Id | Claim Number | Claim Text | Sequence | Dependent |
|---|---|---|---|---|
| 10606734 | 1 | 1. A system for facilitating intelligent mobil… | 0 | |
| 10606734 | 2 | 2. The system of claim 1, wherein the one or … | 1 | claim 1 |
| 10606734 | 3 | 3. The system of claim 1, further comprising … | 2 | claim 1 |
| 10606734 | 4 | 4. The system of claim 3, wherein the automat… | 3 | claim 3 |
| 10606734 | 5 | 5. The system of claim 1, wherein the feature… | 4 | claim 1 |
| 10606734 | 6 | 6. A computer-program product for intelligent … | 5 | |
| 10606734 | 7 | 7. The computer program product of claim 6, w… | 6 | claim 6 |
| 10606734 | 8 | 8. The computer program product of claim 6, f… | 7 | claim 6 |
| 10606734 | 9 | 9. The computer program product of claim 8, w… | 8 | claim 8 |
| 10606734 | 10 | 10. The computer program product of claim 6, … | 9 | claim 6 |

Table 1: The hierarchical structure of a claim

As shown in figure 2, the mean length of claims was 162. In this study, we set max_seq_length parameter in 510 based on BERT input limitation (512 tokens). The SBERT model receives a fixed length of sentence as input. Therefore, we used padding and truncating for the appropriate representations of all input to the maximum length that was 510. In the padding approach, for sentences that are shorter than the maximum length, the model adds empty tokens to the sentences to make up the length.

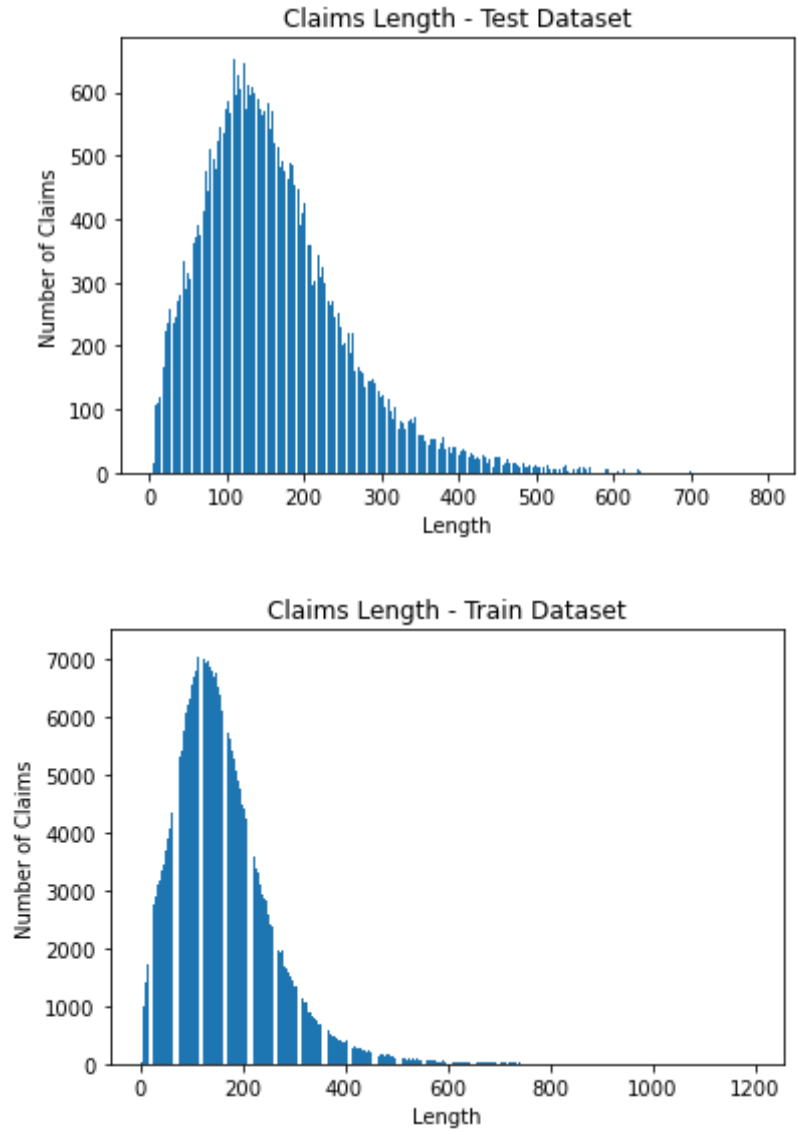

Figure 2: The length of claims

## Patent Classification: CPC Classes

The CPC system and the IPC (International Patent Classification) system are two of the most commonly used classification systems. The CPC is a more specific and detailed version of the IPC system. CPC has a hierarchical structure for classification including Section, Class, Subclass, and Group. At the Subclass level, CPC has 667 labels (Degroote and Held, 2018).

As shown in Figure 3, in the dataset we had 663 labels that 159 of them have less than 350 samples in the dataset. This distribution of labels has led to an imbalance of patent data that KNN is weak for dealing with. Generally, with increasing the number of instances (patents), we can increase the accuracy of the model and the effect of this limitation automatically will be decreased. In this study we used about 1.5M claims, it is clear that by increasing the number of the patents for instance to 3M we can have a better accuracy. For example, when we just used labels that have more than 350 patents (Sofean, 2019), 504 unique main CPC classes (subclass level) were extracted from the dataset. Based on this step, we evaluated the model by 10,000 patents, in this case, F1 Score was increased by 67.23%.

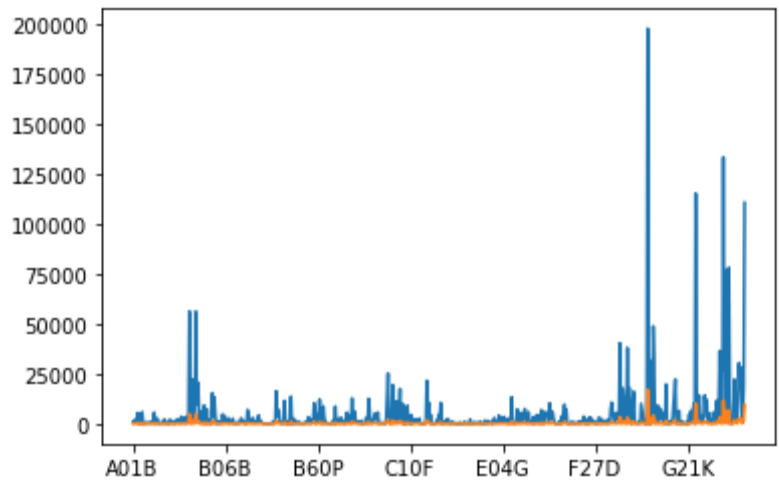

Figure 3: The distribution of patents labels (Training dataset: blue and Test dataset: orange)

## 4. Method and experimental setup

Pretrained Language Models (LMs) have recently become popular in Natural Language Processing (NLP). Pretrained LMs such as ELMo (Peters et al., 2018), BERT (Devlin et al., 2019), and OpenAI GPT (Radford et al., 2018) encode contextual information and high-level features of the language, modeling syntax and semantics, achieving state-of-the-art performance across a variety of tasks, such as named entity recognition (Peters et al., 2017), machine translation (Ramachandran et al., 2017) and text classification (Howard and Ruder, 2018). Since these models are first pre-trained on large corpora, in the next step, fine-tuned on downstream tasks by adding one or more task-specific layers trained from scratch through in-domain data while other layers are frozen.

Generally, these models have a good performance in a particular area of NLP tasks. For example in pairwise sentence semantic similarity, SBERT and BERT are two different approaches that have significant results. While BERT often achieves higher performance (as shown in Table 2), it is too slow for many practical use cases. On the other hand, SBERT has appropriate performance on the practical side, but requires in-domain training data and fine-tuning over the target task to achieve competitive performance (Thakur et al., 2020). In this study, SBERT was used for calculating p2p similarity. The most important advantage of using SBERT is that a practical infrastructure for calculating p2p similarity with the performance of BERT can be proposed.

| Model | Spearman |
|---|---|
| *Not trained for STS* | |
| Avg. GloVe embeddings | 58.02 |
| Avg. BERT embeddings | 46.35 |
| InferSent - GloVe | 68.03 |
| Universal Sentence Encoder | 74.92 |
| SBERT-NLI-base | 77.03 |
| SBERT-NLI-large | 79.23 |
| *Trained on STS benchmark dataset* | |
| BERT-STSb-base | 84.30 ± 0.76 |
| SBERT-STSb-base | 84.67 ± 0.19 |
| SRoBERTa-STSb-base | **84.92 ± 0.34** |
| BERT-STSb-large | **85.64 ± 0.81** |
| SBERT-STSb-large | 84.45 ± 0.43 |
| SRoBERTa-STSb-large | 85.02 ± 0.76 |
| *Trained on NLI data + STS benchmark data* | |
| BERT-NLI-STSb-base | **88.33 ± 0.19** |
| SBERT-NLI-STSb-base | 85.35 ± 0.17 |
| SRoBERTa-NLI-STSb-base | 84.79 ± 0.38 |
| BERT-NLI-STSb-large | **88.77 ± 0.46** |
| SBERT-NLI-STSb-large | 86.10 ± 0.13 |
| SRoBERTa-NLI-STSb-large | 86.15 ± 0.35 |

Table 2: Evaluation BERT and SBERT on the STS benchmark test set (Reimers and Gurevych, 2019)

As shown in Figure 4, in this study, the Augmented SBERT and KNN algorithms were used to predict the class and subclass of patents. In this hybrid approach, the model finds the top N claims, which are higher semantic similarity with the input claim, and in the next step, based on K nearest neighbors; sub-class labels of the patent are predicted.

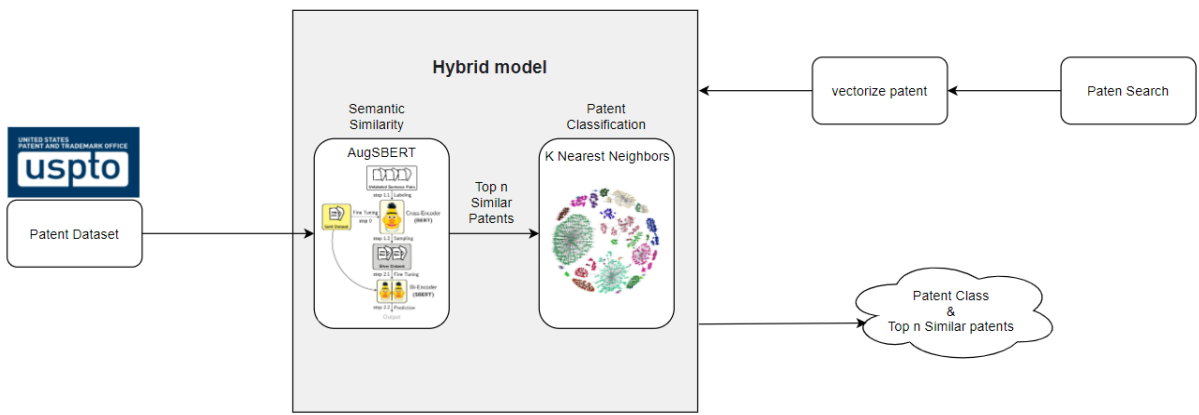

Figure 4: The conceptual representation of the research model

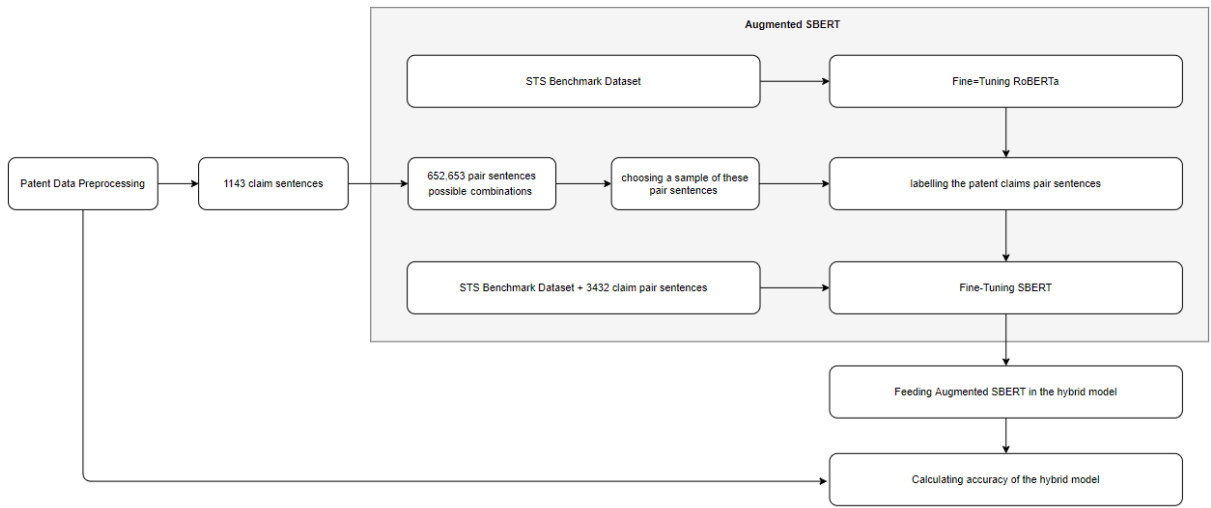

Figure 5: Research process

For the first part of the model, we used the semi-supervised approach as Data Augmentation Method for training SBERT. In this approach, a cross-encoder is used to label a more extensive set of input pairs to augment the training data for fine-tuning the SBERT. Thakur et al. (2020) showed that, in this process, selecting the sentence pairs is crucial for the success of the method. They evaluated their approach on multiple tasks and showed that the augmented approach could improve the performance compared to the original bi-encoder. The process of this approach is illustrated in Figure 6 (Thakur et al., 2020).

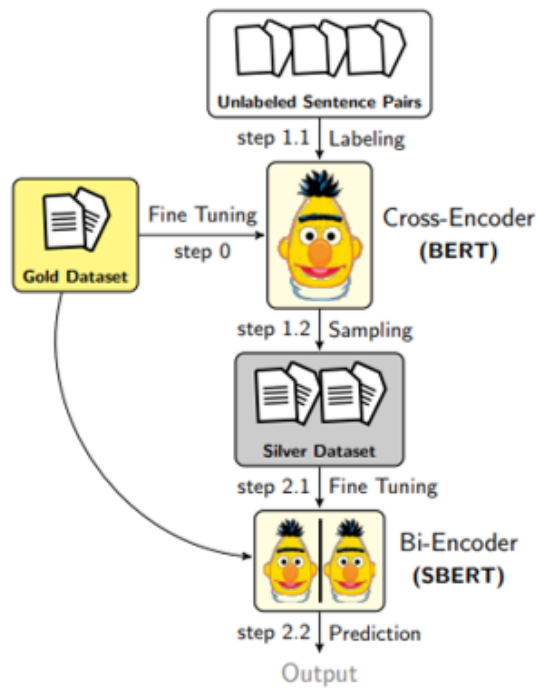

Figure 6: Augmented SBERT In-domain approach (Thakur et al., 2020)

In this study, we have only unlabeled claim-pairs sentences (No annotated datasets). Therefore as shown in Figure 5, for fine-tuning SBERT the Augmented (Domain-Transfer) strategy was chosen. In this approach, in-domain sentence pairs are labeled with the cross-encoder. For n in-domain sentences, there are $n \times (n - 1)/2$ possible combinations. Using all possible combinations would not lead to performance improvement. Therefore, the right sampling strategy helps the model to achieve performance improvement and decrease an excessive computational overhead. In the next step, the bi-encoder is trained on this training dataset (STS Benchmark and In-Domain Labeled dataset). This trained model was called Augmented SBERT (AugSBERT) (Thakur et al., 2020).

The result of Thakur et al. (2020) showed that this approach can achieve an improvement of up to 6 points for in-domain and of up to 37 points for domain adaptation tasks compared to SBERT performance. This approach involves three steps following.
- Step 1: Fine-tuning RoBERTa as a cross-encoder over a small STS benchmark dataset was done.
- Step 2: Using trained RoBERTa to label our patent claims dataset (unlabeled sentence pairs). In this step, we used 1143 claim sentences; there are 652,653 possible combinations based on a particular sampling strategy, 3432 pair claim sentences chosen for fine-tuning SBERT.
- Step 3: Finally, training SBERT on the labeled target dataset (a small STS + 3432 pair claims sentences) was done.

For the prediction part, we used a Sigmoid function after finding similar semantic patents, and based on this layer, the labels of the instance are predicted.

## 5. Results

The results of the research can be categorized into two parts including P2P similarity and semantic search and CPC prediction parts.

### P2P similarity and semantic search

Patent Semantic Search (PSS) is the fundamental part of almost all patent analysis tasks such as technology surveys, technology mapping, and forecasting (Shalaby and Zadrozny, 2018). PSS is related to effective approaches and methods for semantic retrieving relevant patent documents in response to a given search request.

Semantic similarity approaches like transformer models are the novel way for solving the vocabulary mismatch between query terms and relevant patents content in keyword-based search approach (Shalaby and Zadrozny, 2018). As we mentioned before, in this study, AugSBERT was used for creating claims Embeddings, based on these Embeddings the semantic similarity between patents can be calculated through measurement metrics such as Cosine Similarity or Euclidean Distance.

Table 3 illustrates an example of input-claim that the model has calculated the semantic similarity score of them compared to other claims of patents. In this example, the most similar patents to id number 8745119 was found. Examination of the content of the identified cases shows well that the model was able to identify the most similar cases among 1,492,294 patents. It should be noted that in order to evaluate the accuracy of the model, in future studies, we want to calculate the accuracy of our approach using classification labels through Mean Reciprocal Rank (MRR).

| ID | CS | Claim | Labels |
|---|---|---|---|
| 8745119 | **Input Query** | A processor comprising: a cache to store one or more instructions including an instruction to perform a multiplication of a first complex number and a second complex number; a decoder to decode the one or more instructions; a register file including a plurality of registers to store packed data including the first complex number and the second complex number; and one or more execution units to perform the one or more instructions, wherein the one or more execution units to generate a final result to include one or more dot-products in response to processing the instruction, wherein a first dot product is generated by multiplying lower portions of the first and second complex number and a second dot product is generated by multiplying upper portions of the first and second complex number, wherein a first result is generated by subtracting the second dot product from the first dot product, wherein a third dot product is generated by multiplying an upper portion of the first complex number with a lower portion of the second complex number and a fourth dot product is generated by multiplying a lower portion of the first complex number and an upper portion of the second complex number, wherein a second result is generated by adding the third and fourth dot product, wherein the final result is generated based on the first result and the second result. | G06F, G06T |
| 8793299 | 0.92 | A processor comprising: a memory to store one or more instructions including a multiply-add instruction; a register file including a plurality of registers to store packed data including a first complex number and a second complex number; and one or more execution units, wherein the one or more execution units to multiply the first and second complex number in response to performing the multiply-add instruction, wherein the one or more execution units to, | G06F, G06T |
| 8725778 | 0.89 | A processor comprising: a register file to store, in a first register, a first packed data having a first plurality of packed data elements; a cache to store a second packed data having a second plurality of packed data elements corresponding to the first plurality of packed data elements; a decoder to receive a packed multiply-add instruction, specifying a destination operand, a size of the packed data elements, and a source operand; and an execution | G06F, G06T |

| 7 | | unit, responsive to the decoded packed multiply-add instruction, to: multiply corresponding packed data elements from the first packed data and the second packed data to generated a third plurality of corresponding products, add corresponding products from adjacent packed data elements pairwise to generate a fourth plurality of sums, and store the fourth plurality of sums as a third packed data in a register of the register file specified as the destination operand. | |
|---|---|---|---|
| 9535706 | 0.89 | A processor comprising: a register file to store, in a first register, a first packed data having a first plurality of packed data elements; a cache to store a second packed data having a second plurality of packed data elements corresponding to the first plurality of packed data elements; a decoder to receive a packed multiply-add instruction, specifying a destination operand, a size of the packed data elements, and a source operand; and an execution unit, responsive to the decoded packed multiply-add instruction, to: multiply corresponding packed data elements from the first packed data and the second packed data to generated a third plurality of corresponding products, add corresponding products from adjacent packed data elements pairwise to generate a fourth plurality of sums, and store the fourth plurality of sums as a third packed data in a register of the register file specified as the destination operand. | G06F |

```
Runtime of the program is 0.475091934204101156

[[{'corpus_id': 579817, 'score': 0.9270444512367249},
  {'corpus_id': 1023642, 'score': 0.897142231464386},
  {'corpus_id': 658060, 'score': 0.8891114592552185},
  {'corpus_id': 423994, 'score': 0.8863711357116699},
  {'corpus_id': 1094955, 'score': 0.8859415650367737},
  {'corpus_id': 1360514, 'score': 0.8850106000900269},
  {'corpus_id': 395790, 'score': 0.8841106295585632},
  {'corpus_id': 1163806, 'score': 0.8831952810287476},
  {'corpus_id': 1017268, 'score': 0.8831660747528076},
  {'corpus_id': 911515, 'score': 0.8811529874801636},
  {'corpus_id': 76106, 'score': 0.8806372880935669},
  {'corpus_id': 503843, 'score': 0.8792139291763306},
  {'corpus_id': 400042, 'score': 0.8759978345870972},
  {'corpus_id': 960064, 'score': 0.8751339912414551},
  {'corpus_id': 364769, 'score': 0.8745055198669434},
  {'corpus_id': 1186955, 'score': 0.874203622341156},
  {'corpus_id': 503867, 'score': 0.873870313167572},
  {'corpus_id': 823809, 'score': 0.8734570741653442},
  {'corpus_id': 344218, 'score': 0.8726418614387512},
  {'corpus_id': 1343626, 'score': 0.8715025782585144}]]
```

Table 3: Semantic similarity search example

# CPC Prediction

In this study, PatentBERT and DeepPatent are used as baselines to benchmark. In Table 4, results of different approaches including ULMFiT, PatentBERT, and DeepPatent, were shown. The best result for F1 Top 5 score at subclass level in PatentBERT and DeepPatent was less than 45% and 43% respectively. Top-N Accuracy being equal to how often-true class matches with any of the N predictions with higher probability by the classification model. For example, Top 5 accuracy is calculated based on how often the true class matches with any one of the five most probable classes predicted by the model (Hu et al., 2018). On the other hand, in all label prediction, we calculate the effectiveness of all labels in the model accuracy metrics.

The result of the model showed that The F1 score at the sub-class level was 66.48%, and the accuracy of the model was 58% at the same level. Based on the F1 score and accuracy of the model can be concluded at the sub-class level, significantly, the model has better accuracy than previous models.

| Method/Refrences | Text Data | N Patents_Train | N Patents_Test | F1 | Precision | Recall | EVAL | Number_Class |
|---|---|---|---|---|---|---|---|---|
| ULMFiT SVM (Hepburn, 2018) | ALTA+WIPO | 45,150 | 30,100 | 78 | N/A | N/A | Score | 8 labels |
| BiLSTM (Hu et al., 2018) | IPC+CLEF-IP | 90,665 | 2,679 | 64 | N/A | N/A | Top_1 | 96 labels |
| TF-ICF (Lim & Kwon, 2016) | Claim | N/A | 564,793 | N/A | 59 | N/A | Score | 650 labels |
|  | Titles, Abstracts | N/A | 564,793 | N/A | 88 | N/A | Score |  |
| DeepPatent (Li et al., 2018) | IPC+Title+Abstract | 580,546 + 161,551 | 1,350 | N/A | 84 | N/A | Top_1 | 637 labels |
|  | IPC+Title+Abstract | 2,000,147 | 49,900 | N/A | 74 | N/A | Top_1 |  |
|  | IPC+Title+Abstract | 580,546 + 161,551 | 1,350 | 55 | 46 | 75 | Top_1 |  |
|  | IPC+Title+Abstract | 2,000,147 | 49,900 | < 43 | < 35 | < 74 | Top_5 |  |
| PatentBERT (Lee & Hasing, 2020) | IPC+Title+Abstract | 1,950,247 | 49,670 | 65 | 81 | 54 | Top_1 | 656 labels |
|  | IPC+Title+Abstract | 1,950,247 | 49,670 | 45 | 30 | 86 | Top_5 |  |
|  | CPC+Claim | 1,950,247 | 49,670 | 67 | 84 | 55 | Top_1 |  |
|  | CPC+Claim | 1,950,247 | 150,000 | 67 | 84 | 55 | Top_1 |  |
|  | CPC+Claim | 1,950,247 | 150,000 | 81 | N/A | N/A | Score | 8 labels |
| (Hain et al., 2021) | Titles, Abstracts | 1,000,000 | 10,000 | 52 | 54 | 53 | Top_1 | 637 labels |
| PatentSBERTa | CPC+Claim | 1,492,294 | 119,384 | 66.48 | 74 | 60 | Score | 663 labels |
|  | CPC+Claim | 1,492,294 | 119,384 | 82.44 | 79 | 90 | Score | 8 labels |

Table 4: Patent Classification (Lee & Hsiang, 2020; Hain et al., 2020)

As shown in Figure 7, the best result for the F1 score and accuracy had been achieved when we chose 8 nearest neighbors. In this case, Precision and Recall were 74% and 60%, respectively. Also we tested the model for class level, results for F1 score was 82.44%.

It is noted that, in this approach, we can increase the Precision score by increasing the number of K nearest neighbors. As illustrated in Figure 7, when K is 20, precision will be increased to 86%, however, the F1 score and accuracy will decrease to 58% and 46% respectively.

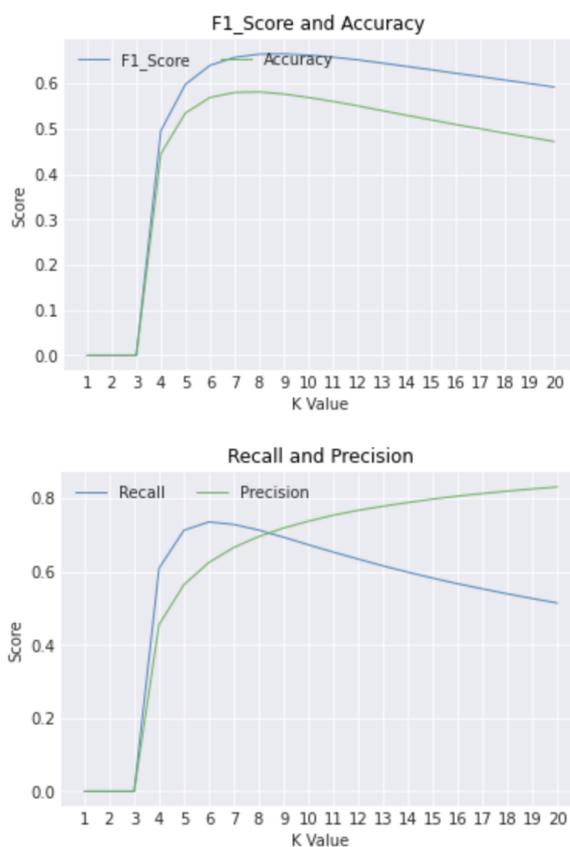

Figure 7: Performance Metrics

Besides the model's accuracy, we focused on proposing the model that can be helpful for the end-user, and results are interpretable and can be compared with existing biological knowledge. This feature can be a significant strength of the model because multi-label classification still is challenging, and using the interpretable model as decision support can help to deploy the model (Zhang & Zhou, 2005). Based on this approach, the model predicts the patent sub-class, and it finds top n the semantic similar patent that the end-user can evaluate the prediction result based on these similar patents.

As pointed out earlier, in this method, for finding the nearest neighbors of a patent in patent classification, semantic similarity based on the SBERT model is used. Indeed, we should calculate and find the most similar patents to input-patents.

## 6. Conclusion

In this paper, we propose an efficient and scalable approach to create transformer-based state-of-the-art embeddings of textual patent data using an augmented SBERT approach. This approach leverages the efficiency of SBERT which makes the creation of embeddings for large or complete patent datasets possible without access to high-end computing infrastructure. Introducing an augmented approach for fine-tuning SBERT, we fine-tuned the general pre-trained SBERT model to the domain of patent claims to increase the model's performance.

A main advantage of the SBERT architecture is the efficiency with embedding distance that can be retrieved. This enables us to create p2p similarity measures for large patent datasets. While the usefulness of text-based p2p similarity measures has been manifold demonstrated for applications such as semantic search, patent landscaping, technology mapping, and patent quality evaluation (cf. e.g. Hain et al., 2020; Arts et al., 2018, 2021), we furthermore demonstrate that our transformer-based p2p similarity can be used for state-of-the-art automated patent classification. Using a simple KNN approach, we are able to predict all assigned patent CPC classes with up to now unrivaled accuracy. We thereby provide a fundamental application of our framework in its own right but also validate that the created embeddings accurately preserve the patent's technological properties. In addition, the simplicity of the CPC prediction model also enables a more understandable and intuitive way to explain, validate, and verify predictions by end-users. Since a patents CPC assignment is a function of the assignment of the K most similar patent, inspecting them makes model decisions understandable and explainable.

## 7. Limitations & Future Research

The size of the sample is the significant limitation of the research that in the future studies, we want to test the model for a larger sample by using Annoy (Approximate Nearest Neighbor Oh Yeah!; Bernhardsson, 2017) and compare the results. The basic idea in Annoy (ANN) is to find approximate similar sentences instead of exact similarity in a space. Using this approach leads to decreasing the time response of the model to query input for the large-scale dataset.

In addition, in future works, we plan to move from using just one claim to multi-input of patent information such as abstracts, descriptions, etc. for calculating p2p semantic similarity and predicting patent multi-label classification, and evaluate our approach based on different inputs. Additionally, we aim at exploring ways for using our approach to a larger length of inputs in order to be applicable to this information without truncating.

Finally, we want to explore approaches for creating a Semantic Textual Similarity (STS) dataset for patents based on an augmented approach and an expert panel. STS assigns a score on the similarity of two texts (Cer et al., 2017). Availability of patent STS dataset can increase the accuracy of transformer models through fine tuning in this field.

## Acknowledgements

This research did not receive any specific grant from funding agencies in the public, commercial, or not-for-profit sectors.